\pdfoutput=1

\documentclass[11pt]{article}

\usepackage[]{emnlp2021}

\usepackage{times}
\usepackage{latexsym}
\usepackage{tablefootnote}

\usepackage[T1]{fontenc}

\usepackage[utf8]{inputenc}

\usepackage{microtype}
\usepackage{times}
\usepackage{latexsym}

\usepackage{microtype}

\usepackage{times}
\usepackage{url}
\usepackage{latexsym}
\usepackage{amssymb}

\usepackage{titlesec}
\usepackage{multirow}
\usepackage{subfig}
\usepackage{mathtools}
\usepackage{colortbl}
\usepackage{float}
\usepackage{xcolor}
\usepackage{enumitem}
\usepackage[capitalize]{cleveref}

\usepackage{booktabs}
\usepackage[font=small,labelfont=bf,tableposition=top]{caption}
\DeclareCaptionLabelFormat{andtable}{#1~#2  \&  \tablename~\thetable}

\title{SgSum:Transforming Multi-document Summarization into Sub-graph Selection}

\author{Moye Chen\thanks{\ \ Equal contribution.}, Wei Li\footnotemark[1], Jiachen Liu, Xinyan Xiao, Hua Wu, Haifeng Wang \\
  Baidu Inc., Beijing, China \\
  \texttt{\{chenmoye,liwei85,liujiachen,xiaoxinyan} \\
  \texttt{wu\_hua,wanghaifeng\}@baidu.com}
  }

\begin{document}
\maketitle
\begin{abstract}
Most of existing extractive multi-document summarization (MDS) methods score each sentence individually and extract salient sentences one by one to compose a summary, which have two main drawbacks: (1) neglecting both the intra and cross-document relations between sentences; (2) neglecting the coherence and conciseness of the whole summary.
In this paper, we propose a novel MDS framework (SgSum) to formulate the MDS task as a \emph{sub-graph selection} problem, in which source documents are regarded as a relation graph of sentences (e.g., similarity graph or discourse graph) and the candidate summaries are its sub-graphs.
Instead of selecting salient sentences, SgSum selects a salient sub-graph from the relation graph as the summary.
Comparing with traditional methods, our method has two main advantages: (1) the relations between sentences are captured by modeling both the graph structure of the whole document set and the candidate sub-graphs;
(2) directly outputs an integrate summary in the form of sub-graph which is more informative and coherent.
Extensive experiments on MultiNews and DUC datasets show that our proposed method brings substantial improvements over several strong baselines.
Human evaluation results also demonstrate that our model can produce significantly more coherent and informative summaries compared with traditional MDS methods.
Moreover, the proposed architecture has strong transfer ability from single to multi-document input, which can reduce the resource bottleneck in MDS tasks.\footnotemark

\footnotetext{Our code and results are available at: \url{https://github.com/PaddlePaddle/Research/tree/master/NLP/EMNLP2021-SgSum}}

\end{abstract}

\section{Introduction}
\label{sub_sec:1}
Currently, most extractive models treat summarization as a sequence labeling task. They score and select sentences one by one \citep{zhong-etal-2020-extractive}.
These models (called sentence-level extractors) do not consider summary as a whole but a combination of independent sentences. This may cause incoherent and redundant problem, and result in a poor summary even if the summary consists of high score sentences. 
Some works \citep{wan2015multi, zhong-etal-2020-extractive} treat summary as a whole unit and try to solve the weakness of sentence-level extractors by using a summary-level extractor. However, these models neglect the intra and cross-document relations between sentences which also have benefits for extracting salient sentences, detecting redundancy and generating overall coherent summaries. 
Relations become more necessary when input source documents are much longer and more complex such as multi-document input.

In this paper, we propose a novel MDS framework called SgSum which formulates the MDS task as a \emph{sub-graph selection} problem.
In our framework, source documents are regarded as a relation graph of sentences (e.g., similarity graph or discourse graph) and the candidate summaries are its sub-graphs. 
In this view, how to generate a good summary becomes how to select a proper sub-graph.
In our framework, the whole graph structure is modeled to help extract salient information from source documents and the sub-graph structures are also modeled to help reflect the quality of candidate summaries.
Moreover, the summary is considered as a whole unit, so SgSum directly outputs the final summary in the form of sub-graph.
By capturing relations between sentences and evaluating summary as a sub-graph, our framework can generate more informative and coherent summaries compared with traditional extractive MDS methods.

\begin{figure*}[t]
    \centering
    \setlength{\abovecaptionskip}{-10mm}
    \includegraphics[scale=0.29]{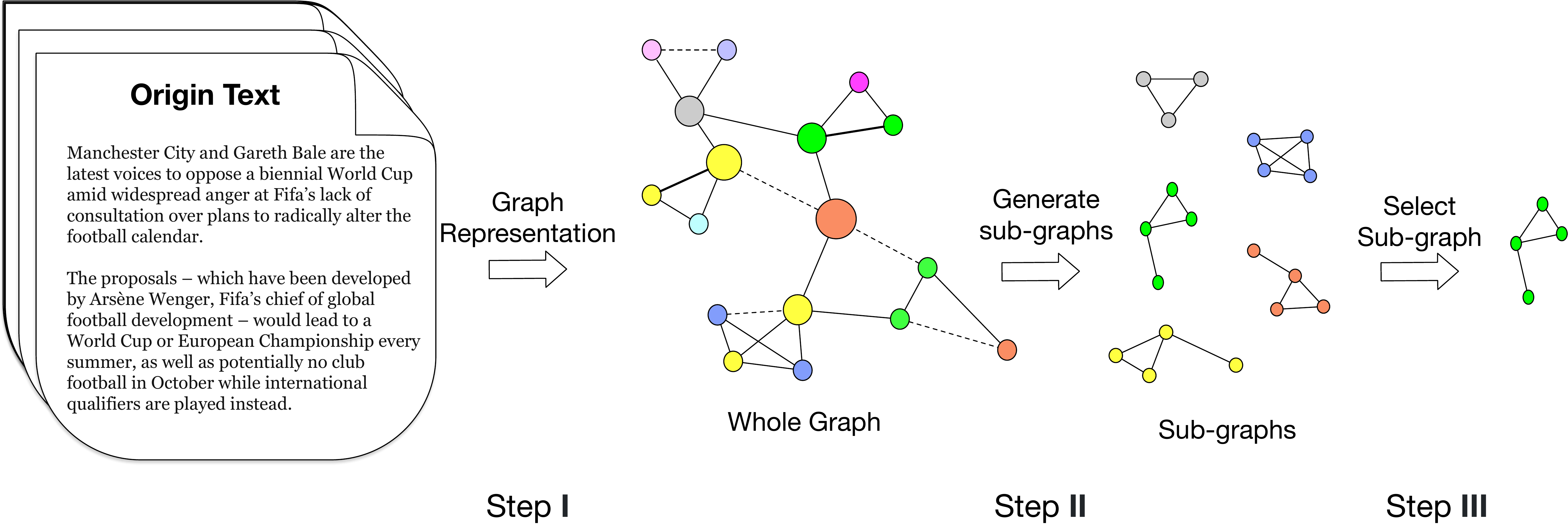}
    \vspace{13mm}
    \captionof{figure}{ 
    Overview of our sub-graph selection framework. Firstly, well-established graph construction methods are used to transform input documents into a graph where sentences are nodes and semantic links between sentences are edges. Then its sub-graphs can be treated as candidate summaries. Finally, we select the best sub-graph as the final summary. 
    }
    \label{fig:1}
    \vspace{-4mm}
\end{figure*}

We evaluate SgSum on two MDS datasets with several types of graphs which all significantly improve the MDS performance. 
Besides, the human evaluation results demonstrate that SgSum can obtain more coherent and informative summaries compared with traditional MDS methods. 
Moreover, the experimental results also indicate that SgSum has strong power on transfer ability when only trained on single-document data. 
It performs much better than several strong MDS baselines including supervised and unsupervised models.

\vspace{1mm}
The contributions of our work are as follows:
\vspace{-1mm}
\begin{itemize}
\setlength{\itemsep}{2pt}
\setlength{\parsep}{0pt}
\setlength{\parskip}{0pt}
    \item We propose a novel framework called SgSum which transforms MDS task into the problem of sub-graph selection. The framework leverages graph to capture relations between sentences, and generates more informative and coherent summaries by modeling sub-graph structures.
    \item Due to the graph-based multi-document encoder, our framework unifies single and multi-document summarization and has strong transfer ability from SDS to MDS task without any parallel MDS training data. Thus, it can reduce the resource bottleneck in MDS tasks.
    \item Our model is general to several well-known graph representations. We experiment with similarity graph, topic graph and discourse graph on two benchmark MDS datasets. Results show that SgSum has achieved superior performance compared with strong baselines.
\end{itemize}

\section{Summarization as Sub-graph Selection}
\label{sub_sec:2}
The graph structure is effective to model relations between sentences which is an essential point to select interrelated summary-worthy sentences in extractive summarization. \citet{erkan2004lexrank} utilize a similarity graph to construct an unsupervised summarization methods called LexRank. G-Flow \citep{christensen2013towards} and DISCOBERT \citep{xu-etal-2020-discourse} both use discourse graphs to generate concise and informative summaries.
\citet{li2016abstractive} and \citet{li2019abstractive} propose to utilize event relation graph to represent documents for MDS.
However, most existing graph-based summarization methods only consider the graph structure of  source document. They neglect that summary is also a graph and its graph structure can reflect the quality of a summary. For example, in a similarity graph, if selected sentences are lexical similar, the summary is probably redundant. And in a discourse graph, if selected sentences have strong discourse connections, the summary tend to be coherent.

We argue that the graph structure of summary is equally important as the source document. Document graph helps to extract salient sentences, while summary graph helps to evluate the quality of summary. Based on this thought, we propose a novel MDS framework SgSum which transforms summarization into the problem of sub-graph selection. SgSum captures relation of sentences both in whole graph structure (source documents) and sub-graph structures (candidate summaries). Moreover, in our framework, summary is viewed as a whole unit in the form of sub-graph. Thus, SgSum can generate more coherent and informative results than traditional sentence-level extractors. 

Figure \ref{fig:1} shows the overview of our framework. Firstly, source documents are transformed into a relation graph by well-known graph construction methods such as similarity graph and discourse graph. 
Sentences are the basic information units and represented as nodes in the graph. And relations between sentences are represented as edges. 
For example, a similarity graph can be built based on cosine similarities between tf-idf representations of sentences. Let $\mathbb{G}$ denotes a graph representation matrix of the input documents, where $\mathbb{G}[i][j]$ indicates the tf-idf weights between sentence $S_i$ and $S_j$. Formally, the task is to generate the summary $S$ of the document collection given $L$ input sentences $S_1$, . . . , $S_L$ and their graph representation  $\mathbb{G}$. 
\vspace{1mm}

\begin{figure}[t]
    \centering
    \includegraphics[scale=0.35]{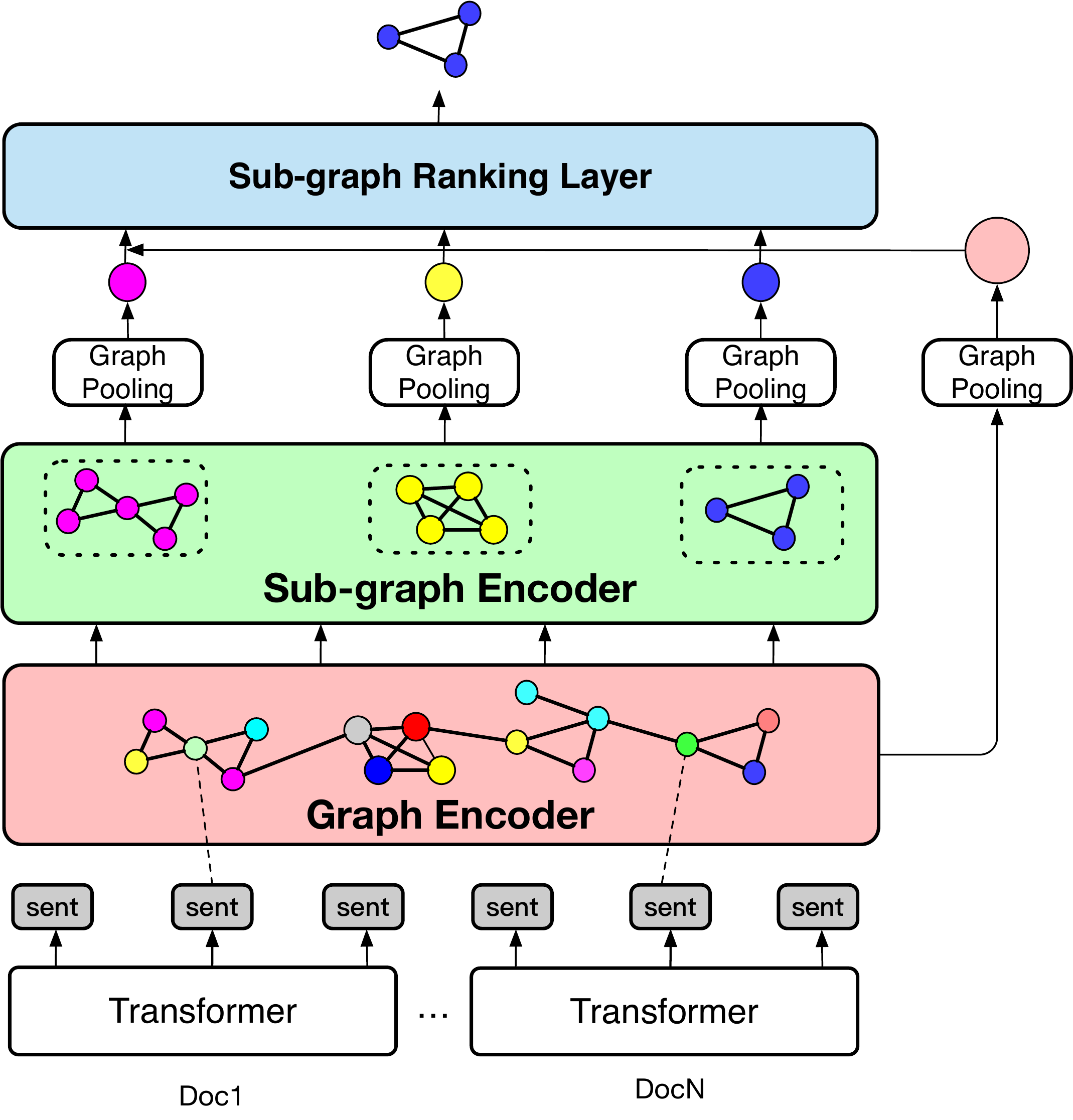}
    \captionof{figure}{ 
    Model architecture of SgSum. Graph-based multi-document encoder takes tokenized documents as input and outputs sentence representations after graph encoding layers. Candidate summaries are modeled by its sub-graph structure in the sub-graph encoder, then scored in a ranking layer.}
    \label{fig:2}
    \vspace{-4mm}
\end{figure}

As Figure \ref{fig:1} shows, if we represent the source documents as a graph, it can be easily observed that sentences will form plenty of different sub-graphs.  By further modelling the sub-graph structures, we can distinguish the quality of different candidate summaries and finally select the best one. Compared with the whole document graph view, sub-graph view is more appropriate to generate a coherent and concise summary. This is also the key point of our framework. Additionally, important sentences usually build up crucial sub-graphs. So it is a simple but efficient way to generate candidate sub-graphs based on those salient sentences.

\section{Methodology}
\subsection{Graph-based Multi-document Encoder}
\label{sub_sec:3.1}
In this section, we introduce our graph-based multi-document encoder. It takes a multi-document set as input and represents all sentences by graph structure. It has three main components: (1) Hierarchical Transformer which processes each document independently and outputs the sentence representations. (2) Graph encoding layer which updates sentence representations by modeling the graph structure of documents. (3) Graph pooling layer which helps to generate an overall representation of source documents. Figure \ref{fig:2} illustrates the overall architecture of SgSum.

\vspace{2mm}
\noindent\textbf{Hierarchical Transformer} \ Most previous works \citep{cao2017improving, jin2020multi, wang-etal-2017-affinity} did not consider the multi-document structure. They simply concatenate all documents together and treat the MDS as a special SDS with longer input.
\citet{wang-etal-2020-heterogeneous} preprocess the multi-document input by truncating lead sentences averagely from each document, then concatenating them together as the MDS input. These preprocessing methods are simple ways to help the model encode multi-document inputs. But they do not make full use of the source document structures. Lead sentences extracted from each document might be similar with each other and result in redundant and incoherent problems.
In this paper, we encode source documents by a Hierarchical Transformer, which consists of several shared-weight single Transformers \citep{vaswani2017attention} that process each document independently. Each Transformer takes a tokenized document as input and outputs its sentence representations. This architecture enables our model to process much longer input. 

\vspace{1mm}
\noindent\textbf{Graph Encoding} \ To effectively capture the relations between sentences in source documents, we incorporate explicit graph representations of documents into the neural encoding process via a graph-informed attention mechanism similar to \citet{li2020leveraging}. Each sentence can collect information from other related sentences to capture global information from the whole input. The graph-informed attention mechanism extends the vanilla self-attention mechanism to consider the pairwise relations in explicit graph representations as:

\vspace{-5mm}
{\fontsize{10}{11}\selectfont
\begin{align}
&\alpha_{ij} = {\rm Softmax}(e_{ij}+R_{ij})
\end{align}
\vspace{-5mm}
}

\noindent where $e_{ij}$ denotes the origin self-attention weights between sentences $S_i$ and $S_j$, $\alpha_{ij}$ denotes the adjusted weights by graph structure. The key point of the graph-based self-attention is the additional pairwise relation bias $R_{ij}$, which is computed as a Gaussian bias of the weights of graph representation matrix $\mathbb{G}$:

\vspace{-5mm}
{\fontsize{10}{11}\selectfont
\begin{align}
&R_{ij} = -\frac{(1-\mathbb{G}_{ij})^2}{2\sigma^2}
\end{align}
\vspace{-5mm}
}

\noindent where $\sigma$ denotes the standard deviation that represents the influence intensity of the graph structure. Then a two-layer feed-forward network with ReLU activation and a high-way layer normalization are applied after the graph-informed attention mechanism. These three components form the graph encoding layers.

\vspace{1mm}
\noindent\textbf{Graph Pooling} \ In the MDS task, information is more massive and relations between sentences are much more complex. So it is necessary to have an overview of the central meaning of multi-document input. 
\citet{zhong-etal-2020-extractive} generate a document representation with Siamese-BERT to guide the training and inference process.
In this paper, based on the graph representation of documents, we apply a multi-head weighted-pooling operation similar to \citet{liu-lapata-2019-hierarchical} to capture the global semantic information of source documents. It takes sentence representations in the source graph as input and outputs an overall representation of them (denoted as $D$), which provides global information of documents for both the sentence and summary selection processes. 

Let $x_i$ denotes the graph representation of sentence $S_i$. For each head $z \in \{1,...,n_{head}\}$, we first transform $x_i$ into attention scores $a_i^z$ and value vectors $b_i^z$, then we calculate an attention distribution $\hat{a}_i^z$ over all sentences in the source graph based on attention scores:

\vspace{-5mm}
{\fontsize{10}{11}\selectfont
\begin{align}
&a_i^z = W_a^zx_i
\end{align}
\vspace{-8mm}
}

\vspace{-5mm}
{\fontsize{10}{11}\selectfont
\begin{align}
&b_i^z = W_b^zx_i
\end{align}
\vspace{-8mm}
}

\vspace{-5mm}
{\fontsize{10}{11}\selectfont
\begin{align}
&\hat{a}_i^z = exp(a_i^z) / \sum_{i=1}^nexp(a_i^z)
\end{align}
\vspace{-5mm}
}

\noindent We next apply a weighted summation with another linear transformation and layer normalization to obtain vector $head_z$ for the source graph. Finally, we concatenate all heads and apply a linear transformation to ontain the global representation $D$:
\vspace{-5mm}
{\fontsize{10}{11}\selectfont
\begin{align}
&head_z = LayerNorm(W_c^z\sum_{i=1}^n\hat{a}_i^zb_i^z)
\end{align}
\vspace{-5mm}
}

\vspace{-5mm}
{\fontsize{10}{11}\selectfont
\begin{align}
D = W_d[head_1||...||head_z]
\end{align}
\vspace{-3mm}
}

\noindent where $W_a^z$, $W_b^z$, $W_c^z$ and $W_d$ are weight matrices, and $||$ denotes the concatenating operator.

Based on the graph-based multi-document encoder, our model can process much longer input than traditional summarization models. Furthermore, our model can treat SDS and MDS as similar tasks in the unified \emph{sub-graph selection} framework.

\subsection{Select from Graph}
\label{sub_sec:3.2}
\textbf{Sub-graph Encoder} \ As we mentioned in Section \ref{sub_sec:2}, sub-graph structure can reflect the quality of candidate summaries. A sub-graph with similar nodes means a redundant summary. And a sub-graph with unconnected nodes represents an incoherent summary. So we apply a sub-graph encoder which has the same architecture with the graph encoder to model each sub-graph. Then we score each sub-graph in a sub-graph ranking layer to select the best sub-graph as the final summary.

\vspace{1mm}
\noindent\textbf{Sub-graph Ranking Layer} \ In the training process, we first calculate ROUGE scores of each sentence with the gold summary. Then we select top-K scoring sentences and make a combination of them to form candidate summaries. The sentences in each candidate summary form a subgraph of the source document graph. 

There are two principles to optimize our framework. Firstly, a good summary can represent the central meaning of source documents which indicates that a good sub-graph should also represent the whole graph. Specifically, the global document representation $D$ which reflects the overall meaning of source documents should be semantic similar with the gold summary. We use a greedy method \citep{nallapati2017summarunner} to extract an oracle summary (composed by source sentences) with the largest ROUGE score corresponding to the abstractive reference summary. Then, sentences in the oracle summary are considered as gold summary sentences, which also form a sub-graph. Let $C^*$ denotes the gold summary and the similarity score between $C^*$ and $D$ is measured by $f(D, C^*)=cosine(D, C*)$, which form the following summary-level loss:

\vspace{-5mm}
{\fontsize{10}{11}\selectfont
\begin{align}
&L_{sum1} = 1 - f(D, C^*)
\end{align}
\vspace{-5mm}
}

Furthermore, we also design a pairwise margin loss for all the candidate summaries similar with \citet{zhong-etal-2020-extractive}. We sort all candidate summaries in descending order of ROUGE scores with the gold summary. All candidate summaries are also represented in the form of sub-graph by using sub-graph encoder. Naturally, the candidate pair with a larger ranking gap should have a larger margin, which is the second principle to design our loss function:

\vspace{-5mm}
{\fontsize{9}{11}\selectfont
\begin{align}
&L_{sum2} = max(0,f(C_j,C^*)-f(C_i,C^*)+\gamma)(i<j)
\end{align}
\vspace{-5mm}
}

\noindent where $C_i$ represents the candidate summary ranked $i$ and $\gamma$ is a hyperparameter used to distinguish between good and bad candidate summaries. $L_{sum1}$ and $L_{sum2}$ compose a summary-level loss function:

\vspace{-6mm}
{\fontsize{10}{11}\selectfont
\begin{align}
&L_{sum} = L_{sum1} + L_{sum2}
\end{align}
\vspace{-6mm}
}

Additionally, we adopt a traditional binary cross-entropy loss between candidate sentences and oracles to learn more accurate sentence and summary representations.

\vspace{-6mm}
{\fontsize{10}{11}\selectfont
\begin{align}
&L_{sent} = -\sum_{i=1}^n(y_i^*\log(\hat{y}_i)+(1-y_i^*)\log(1-\hat{y}_i))
\end{align}
\vspace{-6mm}
}

\noindent where a label $y_i\in\{0, 1\}$ indicates whether the sentence $S_i$ should be a summary sentence. Finally, our loss can be formulated as:

\vspace{-6mm}
{\fontsize{10}{11}\selectfont
\begin{align}
&L = L_{sent} + L_{sum}
\end{align}
\vspace{-6mm}
}

During inference, there are hundreds of sentences in a multi-document set which means there are thousands of sub-graphs need to be considered. In order to overcome this difficulty, we adopt a greedy strategy by first selecting several salient sentences as candidate nodes and then making a combination of them to generate candidate sub-graphs. As the important sentences usually build up crucial sub-graphs, it is a simple way to generate candidate sub-graphs based on those salient sentences. Then we calculate cosine similarities between all sub-graphs with the global document representation $D$ in the sub-graph ranking layer, and select the sub-graph with the highest score as the final summary. Thus, our model can be viewed as a \emph{sub-graph selection} framework which means selecting a proper sub-graph from a whole graph.

Furthermore, the graph structure can help to reorder the sentences in the summary to obtain a more coherent summary \citep{christensen2013towards}. We order the summary by placing sentences with discourse relations next to each other.

\section{Experiments}
\subsection{Experimental Setup}
\label{sub_sec:4.1}

\noindent\textbf{Graph types} \ We experiment with three well-established graph representations: similarity graph, topic graph and discourse graph. (1) The similarity graph is built based on tf-idf cosine similarities between sentences to capture lexical relations. (2) The topic graph is built based on LDA topic model \citep{blei2003latent} to capture topic relations. 
The edge weights are cosine similarities between the topic distributions of sentences. (3) The discourse graph is built to capture discourse relations based on discourse markers (e.g. however, moreover), co-reference and entity links as in \citet{christensen2013towards}. Other types of graphs can also be used in our model. In our experiments, if not explicitly stated, we use the similarity graph by default as it is the most widely used in previous work.

\vspace{-0.1mm}
\noindent\textbf{MultiNews Dataset} \ The MultiNews dataset is a large-scale multi-document summarization dataset introduced by \citep{fabbri-etal-2019-multi}. It contains 56,216 articles-summary pairs and each example consists of 2-10 source documents and a human-written summary. Following their experimental settings, we split the dataset into 44,972/5,622/5,622 for training, validation and testing and truncate each document to 768 tokens.

\noindent\textbf{DUC Dataset} \ We use the benchmark datasets from the Document Understanding Conferences (DUC) containing clusters of English news articles and human reference summaries. We use DUC 2002, 2003 and 2004 datesets which contain 60, 30 and 50 clusters of nearly 10 documents respectively. Four human reference summaries have been created for each document cluster by NIST assessors. Our model is trained on DUC 2002, validated on DUC 2003, and tested on DUC 2004.  We apply the similar preprocessing method with previous work \citep{cho-etal-2019-improving} and truncate each document to 768 tokens

\noindent\textbf{Training Configuration} \ We use the base version of RoBERTa \citep{liu2019roberta} to initialize our models in all experiments. The optimizer is Adam \citep{kingma2014adam} with $\beta1$=0.9 and $\beta2$=0.999, and the learning rate is 0.03 for MultiNews and 0.015 for DUC. We apply learning rate warmup over the first 10000 steps and decay as in \citep{kingma2014adam}. Gradient clipping with maximum gradient norm 2.0 is also utilized during training. All models are trained on 4 GPUs (Tesla V100) for about 10 epochs. We apply dropout with probability 0.1 before all linear layers. The number of hidden units in our models is set as 256, the feed-forward hidden size is 1,024, and the number of heads is 8. The number of transformer encoding layers and graph encoding layers are set as 6 and 2, respectively. As we mentioned in Section \ref{sub_sec:3.2}, during inference we select several salient candidate nodes to build up sub-graphs. And the number of nodes in a sub-graph is determined by the average number of sentences in the gold summary. For MultiNews and DUC, we set the number of candidate nodes and sub-graph nodes as 10/9 and 7/5, respectively.

\subsection{Evaluation Results}
\label{sub_sec:4.2}
We evaluate our models on both the MultiNews and DUC datasets to validate their effectiveness on different types of corpora. The summarization quality is evaluated using ROUGE F1 \citep{lin2004rouge}. We report unigram and bigram overlap (ROUGE-1 and ROUGE-2) between system summaries and gold references as a means of assessing informativeness, and the longest common subsequence (ROUGE-L2) as a means of accessing fluency. 

\noindent\textbf{Results on MultiNews} \ Table \ref{table:1} summarizes the evaluation results on the MultiNews. Several strong extractive and abstractive baselines are evaluated and compared with our models. The first block in the table shows results of extractive methods: LexRank \citep{erkan2004lexrank}, MMR \citep{carbonell1998use}, HeterGraph \citep{wang-etal-2020-heterogeneous} and MatchSum \citep{zhong-etal-2020-extractive} which is the previous extractive SOTA model on the MultiNews dataset. The second block shows results of abstractive methods: PG-MMR \citep{lebanoff-etal-2018-adapting}, Hi-MAP \citep{fabbri-etal-2019-multi}, FT(Flat Transformer) and GraphSum \citep{li2020leveraging}  which is the previous abstractive SOTA model. We report their results following \citet{zhong-etal-2020-extractive, wang-etal-2020-heterogeneous, li2020leveraging}. The last block shows the results of SgSum. Compared with both previous extractive and abstractive SOTA models, SgSum achieves more than 1.1/1.2/0.9 improvements on R-1, R-2 and R-L which demonstrates the effectiveness of our \emph{sub-graph selection} framework.

\begin{table}[t]
\setlength{\tabcolsep}{3.7mm}{
\begin{tabular}{lccc}
\hline
\textbf{Models}&      R-1& R-2& R-L\\
\hline
LexRank&    40.27& 12.63& 37.50\\
MMR&        44.72& 14.92& 40.07\\
MatchSum&   46.20& 16.51& 41.89\\
HeterGraph& 46.05& 16.35& 42.08\\
\hline
PG-MMR&     43.77& 15.38& 39.72\\
Hi-MAP&     44.17& 16.05& 40.35\\
FT&         44.32& 15.11& -\\
GraphSum&   46.07& 17.42& 42.22\\
\hline
SgSum & 47.36& 18.61& 43.13 \\
SgSum(extra)& \textbf{47.53}& \textbf{18.75}& \textbf{43.31} \\
\hline
\end{tabular}
\setlength{\abovecaptionskip}{0pt}%
\setlength{\belowcaptionskip}{5pt}%
\caption{Evaluation results on the MultiNews test set using ROUGE F1\footnotemark. R-1, R-2 and R-L are abbreviations for ROUGE-1, ROUGE-2 and ROUGE-L, respectively.}
\label{table:1}}
\vspace{-4mm}
\end{table}
\footnotetext{-n 2 -m -w 1.2 -c 95 -r 1000 -l 250}

Furthermore, due to our graph representation and graph-based multi-document encoder, our model has the ability to unify single and multi-document summarization task. In our framework, a single document can also be viewed the same as multi-document input. So our model can be enhanced by feeding extra single-document training data. In the last block, \textit{extra} means we leverage CNN/DM data as an extra training resource to improve our model. The results show that single-document data boost the performance of our unified model a further step and achieve a new SOTA result on Multinews.

\vspace{1mm}
\noindent\textbf{Results on DUC} \ Table \ref{table:2} summarizes the evaluation results on the DUC2004 dataset. The first block shows four popular unsupervised baselines, and the second block shows several strong supervised baselines. We report the results of KSLSumm \citep{haghighi-vanderwende-2009-exploring}, LexRank \citep{erkan2004lexrank}, DPP \citep{kulesza2011learning}, Sim-DPP \citep{cho-etal-2019-improving} following \citet{cho-etal-2019-improving}. Besides, we also report the results of SubModular \citep{lin-bilmes-2010-multi}, StructSVM \citep{sipos-etal-2012-large} and PG-MMR \citep{lebanoff-etal-2018-adapting} as strong baselines. The last block shows the results of our models. The results indicate that our model SgSum consistently outperforms most baselines, which further demonstrate the effectiveness of our model on different types of corpora. 

Additionally, we also test the performance of SgSum-extra which add CNN/DM data as a supplement. It is comparable to Sim-DPP baseline which also uses extra CNN/DM data to train a similarity model. And the results again show that single-document data greatly improves the performance of our model.

\begin{table}[t]
\setlength{\tabcolsep}{3.7mm}{
\begin{tabular}{lccc}
\hline
\textbf{Models}&      R-1& R-2& R-L\\
\hline
KLSumm&         31.04& 6.03& -\\
LexRank&        34.44& 7.11& 30.95\\
DPP&            38.10& 9.14& -\\
SubModular&     38.39& 9.58& - \\
\hline
PG-MMR&          36.42& 9.36& - \\
Sim-DPP&        39.35& 10.14& -\\
StructSVM&      39.37& 9.65& 34.52\\
\hline
SgSum &    38.66& 9.73& 34.02 \\
SgSum(extra)&   \textbf{39.41}& \textbf{10.42}& \textbf{35.41} \\
\hline
\end{tabular}
\setlength{\abovecaptionskip}{0pt}%
\setlength{\belowcaptionskip}{5pt}%
\caption{Evaluation results on the DUC2004 test set. We report R-1, R-2 and R-L scores, and follow the ROUGE setting of \citet{cho-etal-2019-improving}.\footnotemark}
\label{table:2}}
\vspace{-1mm}
\end{table}
\footnotetext{-n 2 -m -w 1.2 -c 95 -r 1000 -l 100}
\vspace{-3mm}

\subsection{Transfer Performances}
\label{sub_sec:4.3}
It is commonly known that deep neural networks achieved great improvement on SDS task recently \citep{liu-lapata-2019-text,zhong-etal-2020-extractive,li2018improving,li2018improving2}. However, such supervised models can not work well on MDS task because parallel data for mulit-document are scarce and costly to obtain. For example, the DUC dataset only contains tens of parallel MDS data. There is a pressing need to propose an end-to-end model which is trained on single-document data but can work well with multiple-document input. In this section we do further experiments to verify the transfer ability of our model from single to multi-document task.

We follow the experiment setups of \citet{lebanoff-etal-2018-adapting}, and compare with several strong baseline models: (1) BERTSUMEXT \citep{liu-lapata-2019-text}, an extractive method with pre-trained LM model; (2) PG-MMR \citep{lebanoff-etal-2018-adapting}, an encoder-decoder model which exploits the maximal marginal relevance method to select representative sentences; (3) Extract+Rewrite \citep{song2018structure}, is a recent approach that scores sentences using LexRank and generates a title-like summary for each sentence using an encoder-decoder model. We follow the results of \citet{lebanoff-etal-2018-adapting}. Table \ref{table:3} and Table \ref{table:4} demonstrate the results on MultiNews and DUC2004 respectively. 

\vspace{3mm}
\begin{table}[t]
\setlength{\tabcolsep}{3.1mm}{
\begin{tabular}{lccc}
\hline
\textbf{Models}&      R-1& R-2& R-L\\
\hline
Lead&       40.21& 12.13& 37.13\\
LexRank&   40.27& 12.63& 37.50\\
\hline
BERTSUMEXT& 41.28& 12.05& 37.18 \\
SgSum&      \textbf{43.61}& \textbf{14.07}& \textbf{39.50} \\
\hline
\end{tabular}
\setlength{\abovecaptionskip}{0pt}%
\setlength{\belowcaptionskip}{5pt}%
\caption{Transfer performance on MultiNews dataset}
\label{table:3}}
\end{table}
\vspace{-1mm}

\begin{table}[t]
\setlength{\tabcolsep}{3.1mm}{
\begin{tabular}{lccc}
\hline
\textbf{Models}&      R-1& R-2& R-L\\
\hline
KLSumm&         31.04& 6.03& -\\
LexRank&        34.44& 7.11& 30.95\\
\hline
Extract+Rewrite& 28.90& 5.33& - \\
BERTSUMEXT&    35.13& 8.09& 31.28 \\
PG-MMR&        36.42& 9.36& - \\
SgSum&         \textbf{38.18}& \textbf{9.46}& \textbf{33.81} \\
\hline
\end{tabular}
\setlength{\abovecaptionskip}{0pt}%
\setlength{\belowcaptionskip}{5pt}%
\caption{Transfer performance on DUC2004 dataset}
\vspace{1mm}
\label{table:4}}
\end{table}

\begin{figure}[t]
    \centering
    \includegraphics[scale=0.32]{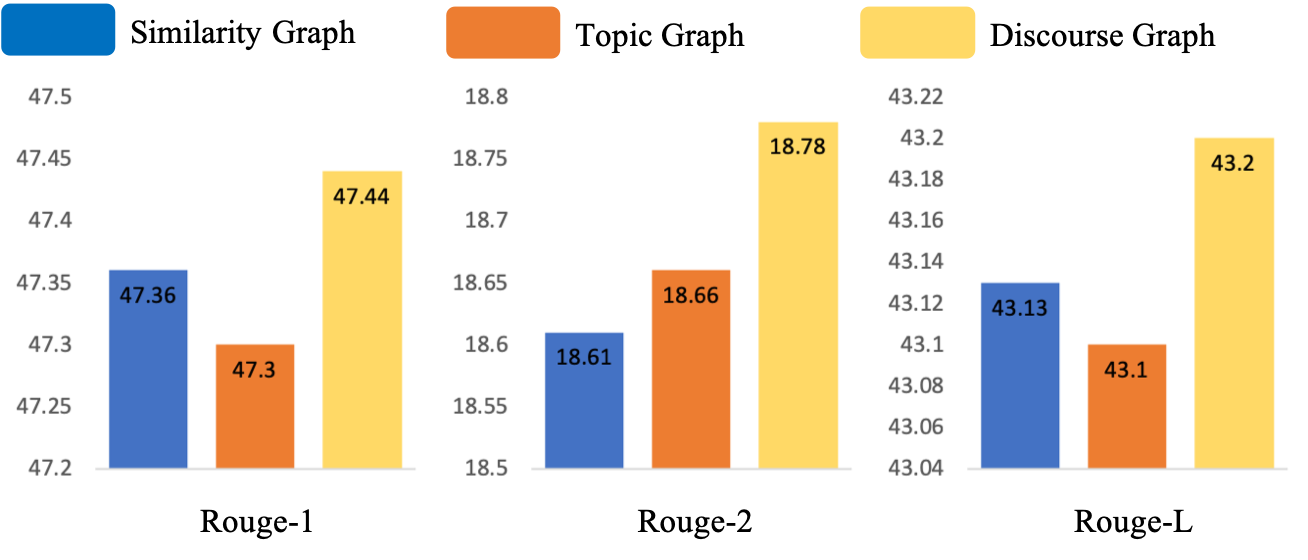}
    \vspace{-3mm}
    \captionof{figure}{ 
    Results on different graph types.}
    \label{fig:3}
    \vspace{-3mm}
\end{figure}

As shown in Tables \ref{table:3} and \ref{table:4}, the second blocks are transfer models which are only trained on SDS data and tested on MDS data directly. BERTSUMEXT, PG-MMR, SgSum are trained on CNN/DM, while Extract+Rewrite is trained on Gigaword.
The results show that our model achieves better performance than several strong unsupervised models. Furthermore, when trained only on the SDS data, SgSum performs much better on transfer ability compared with the three baselines in the second block of Table \ref{table:4}. The above evaluation results on MultiNews and DUC datasets both validate the effectiveness of our model. The sub-graph selection framework greatly improves the performance of MDS and shows a powerful transfer ability which can reduce the resource bottleneck in MDS.

\subsection{Analysis}
\label{sub_sec:4.4}
We further analyze the effects of graph types on our model and validate the effectiveness of different components of our model by ablation studies.

\begin{table}[t]
\setlength{\tabcolsep}{2mm}{
\begin{tabular}{lccc}
\hline
\textbf{Models}&      R-1& R-2& R-L\\
\hline
SgSum&        \textbf{47.36}& \textbf{18.61}& \textbf{43.13}\\
w/o s.g. enc&      46.87& 17.93& 42.67\\
w/o s.g. rank &            46.91& 17.97& 42.80\\
w/o s.g. enc\&rank &           46.69& 17.64& 42.48\\
w/o graph enc&            46.21& 17.12& 42.11\\
w/o all &            45.43& 16.62& 41.32\\
\hline
\end{tabular}
\setlength{\abovecaptionskip}{0pt}%
\setlength{\belowcaptionskip}{5pt}%
\caption{Ablation study on the MultiNews test set. s.g. is the abbreviation for sub-graph.}
\label{table:5}}
\end{table}

\vspace{1mm}
\noindent\textbf{Effects of Graph types} \ We compare the results of similarity graph, topic graph and discourse graph on the MultiNews test set. The comparison results in Figure \ref{fig:3} show that the discourse graph achieves the best performance on all metrics, which demonstrate that graphs with richer relations are more helpful for MDS.

\begin{table*}[!htbp]
\centering
\setlength{\tabcolsep}{2mm}{
\begin{tabular}{l|ccccc|ccccc}
\hline
\multirow{2}{*}{\textbf{Models}}& \multicolumn{5}{c|}{\textbf{Informativeness}}& \multicolumn{5}{c}{\textbf{Coherence}} \\
\cline{2-11}
& \textbf{1st} & \textbf{2nd} & \textbf{3rd} & \textbf{4th} & \textbf{rating} & \textbf{1st} & \textbf{2nd} & \textbf{3rd} & \textbf{4th} & \textbf{rating} \\

\hline
LexRank&    0.13& 0.14& 0.17& 0.56& -0.89$^*$& 0.09& 0.15& 0.11& 0.65& -1.08$^*$ \\
Submodular&     0.29& 0.27& 0.30& 0.14& 0.27$^*$& 0.31& 0.32& 0.26& 0.11& 0.46$^*$\\
BERTSUMEXT&     0.24& 0.30& 0.29& 0.17& 0.15$^*$& 0.23& 0.22& 0.41& 0.14& -0.01$^*$\\
SgSum&          \textbf{0.34}& 0.29& 0.24& 0.13& \textbf{0.47}& \textbf{0.37}& 0.31& 0.22& 0.10& \textbf{0.63}\\
\hline
\end{tabular}
\setlength{\abovecaptionskip}{0pt}%
\setlength{\belowcaptionskip}{5pt}%
\caption{Human evaluation of system summaries on DUC-04. 1st is the best and 4th is the worst. The larger rating denotes better summary quality. $^*$ indicates the overall ratings of the corresponding model are significantly (by Welch’s t-test with $p<0.05$) outperformed by our model. The inter-annotator agreement score (Cohen Kappa) is 0.67, which indicates substantial agreement between annotators.}
\label{table:6}}
\vspace{-5mm}
\end{table*}

\vspace{1mm}
\noindent\textbf{Ablation Study} \ Table \ref{table:5} summarizes the results of ablation studies, which aim to validate the effectiveness of each individual component of our model. ``w/o graph enc'' denotes removing the graph-based multi-document encoder, encoding the source input by concatenating all documents as a sequence. ``w/o subgraph enc'' and ``w/o subgraph rank'' denontes removing the subgraph encoder and the subgraph ranking layer, respectively. ``w/o all'' denotes removing all graph components, which is actually the BERTSUMEXT baseline model.
The experimental results confirmed that our framework which transforms MDS task into sub-graph selection is effective (see w/o subgraph enc and subgraph rank). Besides, incorporating explicit graph structure (see w/o graph enc) also help to process long input source and result in better performances for MDS.

\subsection{Human Evaluation}
\label{sub_sec:4.5}
In addition to the automatic evaluation, we also assess system performance by human evaluation. We use the DUC2004 as human evaluation set, and invite 2 annotators to assess the outputs of different models independently. We use Cohen Kappa \citep{Cohen1960coefficient} to calculate the inter-annotator agreement between annotators. Annotators assess the overall quality of summaries by ranking them considering the following criteria: (1) \textbf{Informativeness}: is the main meaning expressed in the source documents preserved in the summary? (2) \textbf{Coherence}: is the summary coherent between sentences and well-formed? Annotators were asked to ranking all systems from 1 (best) to 4 (worst). All systems get score 2, 1, -1, -2 for ranking 1, 2, 3, 4 respectively. The rating of each system is computed by averaging the scores on all test instances.

Four system summaries are presented in Table \ref{table:6}. The results demonstrate that SgSum is rated as the best on both informativeness and coherence. Regarding the overall ratings, the summaries generated by SgSum are frequently ranked as the best, which significantly outperforms other models. The human evaluation results further validate the effectiveness of our proposed \emph{sub-graph selection} framework.

\section{Related Work}
\subsection{Graph-based Summarization}
\label{sub_sec:5.1}
Most previous graph extractive MDS approaches aim to extract salient textual units from documents based on graph structure representations of sentences.
\citet{erkan2004lexrank} introduce LexRank to compute sentence importance based on the eigenvector centrality in the connectivity graph of inter-sentence cosine similarity.  \citet{christensen2013towards} build multi-document graphs to identify pairwise ordering constraints over the sentences by accounting for discourse relationships between sentences. More recently, \citet{yasunaga-etal-2017-graph} build on the approximate discourse graph model and account for macro-level features in sentences to improve sentence salience prediction.  \citet{yin2019graph} also propose a graph-based neural sentence ordering model, which utilizes an entity linking graph to capture the global dependencies between sentences. \citet{li2020leveraging} incorporate explicit graph representations to the neural architecture based on a novel graph-informed self-attention mechanism. It is the first work to effectively combine graph structures with abstractive MDS model. \citet{wu2021bass} present BASS, a novel framework for Boosting Abstractive Summarization based on a unified Semantic graph, which aggregates co-referent phrases distributing across a long range of context and conveys rich relations between phrases. However, these works only consider the graph structure of source documents, but neglect the graph structures of summaries which are also important to generate coherent and informative summaries.

\subsection{Sentence or Summary-level Extraction}
\label{sub_sec:5.2}
Extractive summarization methods usually produce a summary by selecting some original sentences in the document set by a sentence-level extractor. Early models employ rule-based methods to score and select sentenecs \citep{lin2002single,lin2011class,takamura2009text,schilder-kondadadi-2008-fastsum}. Recently, SUMMARUNNER \citep{nallapati2017summarunner} adopt an encoder based on Recurrent Neural Networks which is the earliest neural summarization model. SUMO \citep{liu2019single} capitalizes on the notion of structured attention to induce a multi-root dependency tree representation of the document. However, all these models belong to sentence-level extractors which select high score sentences individually and might raise redundancy \citep{narayan-etal-2018-ranking}.

Different from above studies, some work focus on the summary-level selection. \citet{wan2015multi} optimize the summarization performance directly based on the characteristics of summaries and rank summaries directly during inference. \citet{bae-etal-2019-summary}, \citet{paulus2017deep}
and \citet{celikyilmaz-etal-2018-deep} use reinforcement learning to globally optimize summary-level performance. Recent studies \citep{alyguliyev2009two, galanis2010extractive, zhang-etal-2019-pretraining} have attempted to a build two-stage document summarization. The first stage is usually to extract some fragments of the original text, and the second stage is to select or modify on the basis of these fragments. \citet{mendes-etal-2019-jointly} follow the extract-then-compress paradigm to train an extractor for content selection. \citet{zhong-etal-2020-extractive} propose a novel extract-then-match framework which employs a sentence extractor to prune unnecessary information, then outputs a summary by matching models.
These methods consider summary as a whole rather than individual sentences. However, they neglect the relations between sentences during both scoring and selecting.

\subsection{From Single to Multi-document}
\label{sub_sec:5.3}
Recent neural network summarization models focus on SDS due to the large parallel datasets automatically harvested from online news websites including Gigaword \citep{rush2017neural}, CNN/DM \citep{hermann2015teaching}, NYT \citep{Sandhaus2008NYT} and Newsroom \citep{grusky-etal-2018-newsroom}. 
However, MDS has not yet fully benefited from the development of neural network models, because parallel data for MDS are scarce and costly to obtain.

A promising route to generating summary from a multi-document input is to apply a model trained for SDS to a “mega-document” \citep{lebanoff-etal-2018-adapting} created by concatenating all documents together. Nonetheless, such a model may not suit well for two reasons. First, identifying important text pieces from a mega-document can be challenging for the model, which is trained on single-document data where the summary-worthy content is often contained in the first few sentences. This is not the case for a mega-document. Second, redundant text pieces in a mega-document can be repeatedly used for summary generation under the current framework. \citet{lebanoff-etal-2018-adapting} present a novel adaptation model, named PG-MMR, to generate summary from multi-document inputs. However, it still considers MDS data as a meta-document.
In contrast, our model unifies SDS and MDS by graph representations, and achieves great performance on transferring from SDS to MDS.

\section*{Conclusion}
\label{sub_sec:6}
We propose a novel framework SgSum which transforms the MDS task into the problem of sub-graph selection. 
SgSum captures the relations between sentences by modelling both the graph structure of the whole document set and the candidate sub-graphs, then directly output an integrate summary in the form of sub-graph which is more informative and coherent.
Experimental results on two MDS datasets show that SgSum brings substantial improvements over several strong baselines.
Moreover, the proposed architecture has strong transfer ability from single to multi-document, which can reduce the resource bottleneck in MDS tasks.

\section*{Acknowledgments}
\label{sub_sec:7}
This work was supported in part by the National Key R\&D Program of China under Grant 2020YFB1406701.

\bibliography{anthology,custom}
\bibliographystyle{acl_natbib}

\appendix


\end{document}